# Fuzzy Mnesors

Gilles CHAMPENOIS

*Collège Saint-André, Saint-Maur, France*

*gilles_champenois@yahoo.fr*

ABSTRACT. A fuzzy mnesor space is a semimodule over the positive real numbers. It can be used as theoretical framework for fuzzy sets. Hence we can prove a great number of properties for fuzzy sets without refering to the membership functions.

> "Classical logic has erred in devoting so little attention to approximate reasoning and focusing to such a high degree on exact reasoning. So when you take a course in logic, you learn all kinds of things which are of very little use in everyday life."
>
> *Lotfi Zadeh*

## I. INTRODUCTION

According to Zadeh's fuzzy theory [1], $k$-16 ($k$-$i$ here represents the number of years a student attend school before leaving) falls under the fuzzy set of *HIGH – EDUCATED* with a membership value of 1.

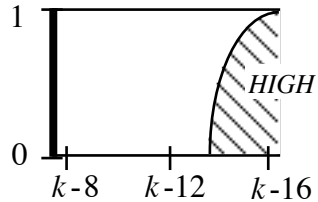

FIGURE 1. The membership function of the fuzzy set *HIGH*

We can now imagine an operator on any fuzzy set which we will call external multiplication: multiplying a fuzzy set $A$ by the positive real number $\lambda$ returns the fuzzy set (written $A\lambda$) whose membership function is:

$$\mu_{A\lambda}(\cdot) = \left(\mu_A(\cdot)\right)^{\lambda^{-1}}$$

To the previous example we add the fuzzy set *HIGH* $\lambda$ with $\lambda = \dfrac{1}{2}$:

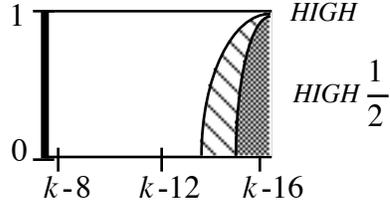

FIGURE 2. The membership functions of the fuzzy sets *HIGH* and $HIGH\frac{1}{2}$

You can see that $HIGH\frac{1}{2}$ is more selective than *HIGH*.

We want here to embed fuzzy set theory in a larger framework called the mnesor theory and based on the semimodule over the positive real numbers.

## II. SEMIMODULE OVER THE POSITIVE REAL NUMBERS

The set of the positive real numbers (denoted $R^+$) is considered with two operations: the multiplication and the maximum operator (written $\oplus$) which will be called addition: for any $x, y \in R^+$, $x \oplus y = \max(x, y)$.

Let $M$ denote a commutative monoid with the identity element $\bar{0}$. We add an external multiplication returning $A\lambda \in M$ from $A \in M$ and $\lambda \in R^+$. $M$ is called a semimodule over $R^+$ iff the four next properties hold for any $\lambda, \mu \in R^+$ and $A, B \in M$:

(1) $A1 = A$
(2) $A\lambda + A\mu = A(\lambda \oplus \mu)$
(3) $(A + B)\lambda = A\lambda + B\lambda$
(4) $(A\lambda)\mu = A(\lambda \times \mu)$

We first derive some basic properties.

*Idempotence*. The addition of fuzzy mnesors is idempotent.

PROOF. For any $A \in M$, $A + A = A1 + A1 = A(1 \oplus 1) = A1 = A$

*Ordering*. An order relation is naturally defined by the addition: $A \subseteq B$ iff $A + B = B$.

PROOF. $\subseteq$ is an order relation indeed, since $M$ is a commutative idempotent monoid.

Note that if $B = A\lambda$ with $\lambda \leq 1$, then $B$ is inferior to $A$ ($A + A\lambda = A1 + A\lambda = A(1 \oplus \lambda) = A1 = A$) and if $B = A\lambda$ with $\lambda \geq 1$, then $B$ is superior to $A$. In the fomer case, we say that $B$ is more selective than $A$ (see the example of $HIGH\frac{1}{2}$).

*Positivity*. All mnesors are positive ($A + \bar{0} = A$) and $M$ is zero-sum-free ($A + B = \bar{0}$ implies that $A = B = \bar{0}$)

PROOF. $A = A + \bar{0} = A + A + B = A + B = \bar{0}$

*Empty mnesor*. $\bar{0}\lambda = \bar{0}$ for all $0 \le \lambda \le 1$

PROOF. $A + \bar{0}\lambda = A + \bar{0} + \bar{0}\lambda = A + \bar{0}1 + \bar{0}\lambda = A + \bar{0}(1 \oplus \lambda) = A + \bar{0}1 = A + \bar{0} = A$, for any $A \in M$. Thus $\bar{0}\lambda$ is itself the identity element.

## III. COMPLEMENT

We suppose now that the top element of $M$ exists (written $\bar{1}$ and called full mnesor) and we add an unary operator on $M$ called complement ($\bar{A}$ is the compement of $A$) that satisfies the following properties:

(5) $\bar{\bar{A}} = A$
(6) $\bar{\bar{1}} = \bar{0}$
(7) $\overline{A\lambda} = \bar{A}\lambda^{-1}$
(8) $\bar{A} \subseteq \bar{B}$ iff $B \subseteq A$

Note that $\bar{1}\lambda = \bar{1}$ for all numbers $\lambda \ge 1$ ($\bar{1}\lambda = \overline{\bar{0}\lambda} = \overline{\bar{0}\lambda^{-1}} = \overline{\bar{0}} = \bar{1}$).

*Intersection*. We define the intersection of the two mnesors $A,B$ by $A \circ B = \overline{\bar{A} + \bar{B}}$ and we prove that the external multiplication distributes over the intersection.

PROOF. $(A\lambda) \circ (B\lambda) = \overline{\overline{A\lambda} + \overline{B\lambda}} = \overline{\bar{A}\lambda^{-1} + \bar{B}\lambda^{-1}} = \overline{(\bar{A} + \bar{B})\lambda^{-1}} = \overline{(\bar{A} + \bar{B})}\lambda = (A \circ B)\lambda$

The intersection is idempotent ($A \circ A = \overline{\bar{A} + \bar{A}} = \bar{\bar{A}} = A$)

*Absorption*. $A \circ B$ is inferior to $A$ and $B$ ($A \circ B \subseteq A$ since $\overline{A \circ B} = \bar{A} + \bar{B} \supseteq \bar{A}$). Thus $A + (A \circ B) = A$. Then by complementing we get $\overline{A + (A \circ B)} = \bar{A}$ or $\bar{A} \circ (\bar{A} + \bar{B}) = \bar{A}$.

*Lattice*. $(M, +, \circ)$ is a lattice.

PROOF. $+, \circ$ are commutative, associative and idempotent. Absorption holds.

# IV. INTERPRETATION OF FUZZY SETS

We want to prove that fuzzy sets satisfy the definition of mnesors. But we first substitute the function $c_k(x) = e^{\frac{k}{\ln(x)}}$ for $x \to 1-x$ in the definition of the complement. Like $1-x$, $c_k(x)$ takes values in $[0\,1]$, decreases monotonously, exchanges 0 and 1 (that is, $c_k(0) = 1$ and $c_k(1) = 0$) and is involutive. Moreover $c_k(x)$ values are very close to $1-x$ if $k \approx 0,4$ as you can see below.

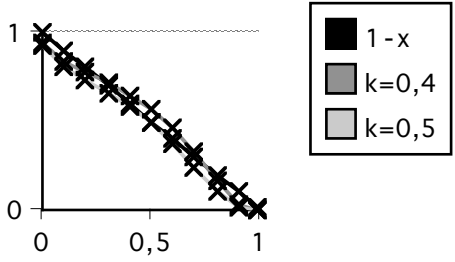

FIGURE 3. $1-x$ function, $c_{0,4}(x)$ and $c_{0,5}(x)$ functions

Note that $(c_k(x))^n = c_{k \times n}(x)$ and $c_k(x^n) = c_{\frac{k}{n}}(x)$.

The reunion of fuzzy sets and the external multiplication satisfy the linear properties and the complement properties.

PROOF. (1) $\mu_{A1}(\cdot) = (\mu_A(\cdot))^{1^{-1}} = (\mu_A(\cdot))^1 = \mu_A(\cdot)$

(2) $\mu_{(A \cup B)\lambda}(\cdot) = (\mu_{A \cup B}(\cdot))^{\lambda^{-1}} = (\mu_A(\cdot) \oplus \mu_B(\cdot))^{\lambda^{-1}} = (\mu_A(\cdot))^{\lambda^{-1}} \oplus (\mu_B(\cdot))^{\lambda^{-1}} = \mu_{A\lambda}(\cdot) \oplus \mu_{B\lambda}(\cdot)$

(3) $\mu_{A(\lambda \oplus \delta)}(\cdot) = (\mu_A(\cdot))^{(\lambda \oplus \delta)^{-1}} = (\mu_A(\cdot))^{\lambda^{-1}} \oplus (\mu_A(\cdot))^{\delta^{-1}} = \mu_{A\lambda}(\cdot) \oplus \mu_{A\delta}(\cdot)$

(4) $\mu_{(A\lambda)\delta}(\cdot) = (\mu_{A\lambda}(\cdot))^{\delta^{-1}} = ((\mu_A(\cdot))^{\lambda^{-1}})^{\delta^{-1}} = (\mu_A(\cdot))^{(\lambda \otimes \delta)^{-1}} = \mu_{A(\lambda \otimes \delta)}(\cdot)$

(5) $\mu_{\overline{\overline{A}}}(\cdot) = \mu_A(\cdot)$ because $c_k$ is involutive

(6) $\mu_{\overline{1}}(\cdot) = c_k(\mu_1(\cdot)) = c_k(1) = 0 = \mu_0(\cdot)$

(7) $\mu_{\overline{A\lambda}}(\cdot) = c_k(\mu_{A\lambda}(\cdot)) = c_k((\mu_A(\cdot))^{\lambda^{-1}}) = c_{k \times \lambda}(\mu_A(\cdot)) = (c_k(\mu_A(\cdot)))^\lambda = (\mu_{\overline{A}}(\cdot))^\lambda = \mu_{\overline{A}\lambda^{-1}}(\cdot)$

(8) If $\mu_A(\cdot) \leq \mu_B(\cdot)$, then $c_k(\mu_A(\cdot)) \geq c_k(\mu_B(\cdot))$, since the $c_k$ function decreases. Hence $\mu_{\overline{A}}(\cdot) \geq \mu_{\overline{B}}(\cdot)$

# REFERENCES

1. ZADEH L. (1965), *Fuzzy Sets*